\def\year{2020}\relax
\documentclass[letterpaper]{article} 
\usepackage{aaai20}  
\usepackage{times}  
\usepackage{helvet} 
\usepackage{courier}  
\usepackage[hyphens]{url}  
\usepackage{graphicx} 
\usepackage{xcolor}
\nocopyright
\title{KnowBias: Detecting Political Polarity in Long Text Content}
\author{\Large \textbf{Aditya Saligrama} 
\\
MIT PRIMES/Weston High School\\
77 Vassar Street\\
Cambridge, MA 02139\\
saligrama@csail.mit.edu 
}
 
\begin{document}

\maketitle

\begin{abstract}
We introduce a classification scheme for detecting political bias in long text content such as newspaper opinion articles. Obtaining long text data and annotations at sufficient scale for training is difficult, but it is relatively easy to extract political polarity from tweets through their authorship. We train on tweets and perform inference on articles. Universal sentence encoders and other existing methods that aim to address this domain-adaptation scenario deliver inaccurate and inconsistent predictions on articles, which we show is due to a difference in opinion concentration between tweets and articles. We propose a two-step classification scheme that uses a neutral detector trained on tweets to remove neutral sentences from articles in order to align opinion concentration and therefore improve accuracy on that domain. Our implementation is available for public use at \texttt{https://knowbias.ml}.
\end{abstract}

\section{Introduction}
Rising bias in news media, along with the formation of filter bubbles on social media, where content with the same political slant is repeatedly shared, have contributed to severe partisanship in the American political environment in recent years \cite{renka_2010,kelly_francois_2018}. We aim to increase awareness of this heightened polarization by alerting users to the political bias in the content they consume.

In this work, we discuss an NLP-based approach that predicts political bias on a left-to-right spectrum on long text such as news articles independent of metadata such as content origin or authorship. Annotating polarity on long documents at sufficient scale for training is infeasible, requiring humans to read each article and manually determine polarity. However, tweets can be easily gathered in high volume and be annotated based on authorship.

We envision an approach that transfers knowledge from tweets to long text at test time. While tweets have been analyzed for political sentiment \cite{demszky2019analyzing}, no research has focused on domain adaptation from short to long text in this context. Previous work has filtered text in order to derive justifiable predictions \cite{lei2016rationalizing}, but not for domain adaptation for our target problem. Universal sentence encoders \cite{2018arXiv180311175C} provide good text representations regardless of target task. We would expect a classifier trained on these to perform well on all text, but this delivers inaccurate and inconsistent predictions. 

We show that this poor performance is due to the existence of neutral, apolitical sentences in articles that dilute opinion concentration compared to tweets. Our proposed method (Figure \ref{pipeline}) alleviates this issue by using a neutral detector trained on tweets to remove neutral sentences before predicting bias, improving prediction accuracy and consistency. Our work summarizes \citeauthor{2019arXiv190500724S} (\citeyear{2019arXiv190500724S}).

\section{Predicting Polarity in Text Content}

\begin{figure}[t!]
    \includegraphics[width=0.45\textwidth]{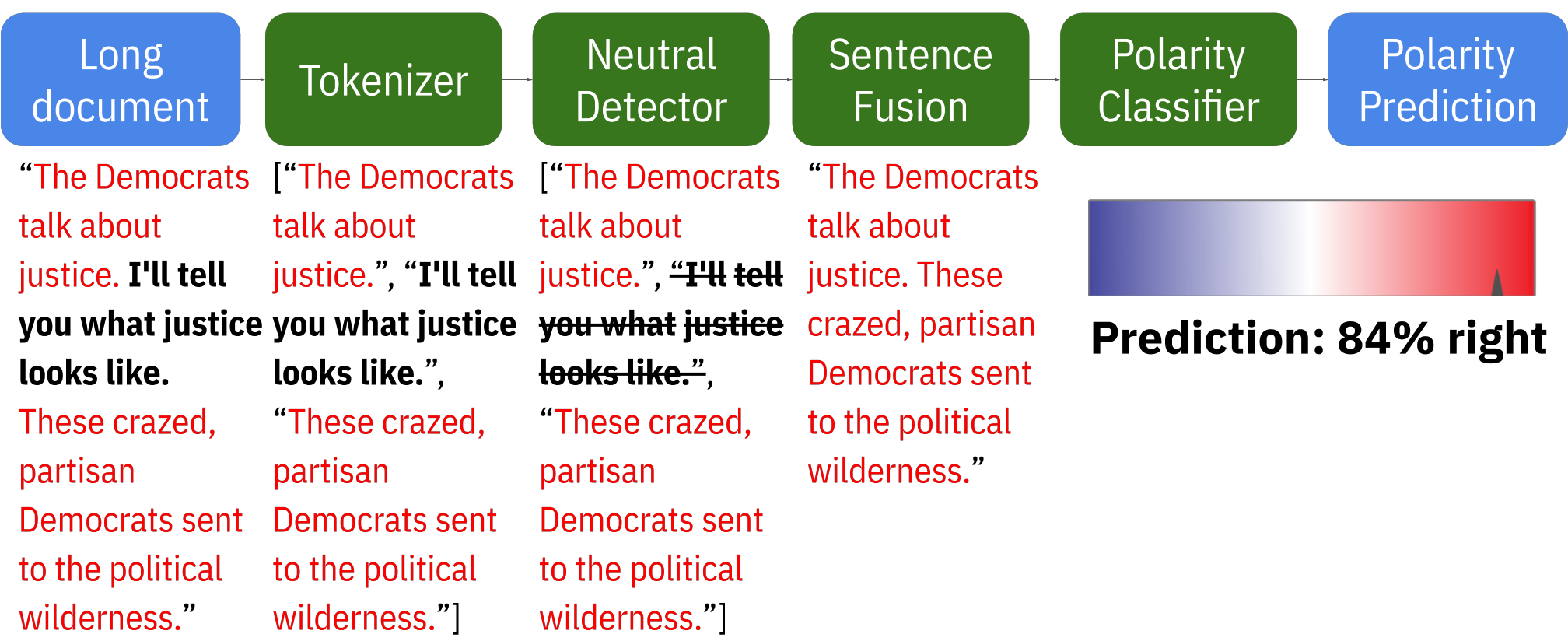}
    \caption{\small Proposed two-step classification scheme that tokenizes sentences in long documents and uses a neutral detector to filter out neutral sentences. Subsequently, it fuses remaining sentences to make a final prediction via a polarity classifier. Red sentences are polarized; black bold sentences are removed by the neutral detector.}
    \label{pipeline}
\end{figure}

\textbf{Data collection} \indent We train on political tweets due to the aforementioned ease in collecting and annotating them at scale and aim to transfer this knowledge to longer articles. Our polarity data consisted of roughly 150,000 tweets from 28 Twitter verified politicians or media personalities across the political spectrum. 80\% of these samples were used for training and 20\% were used as testing. We also sampled a set of roughly 80,000 neutral tweets from the Twitter general stream in order to train the neutral detector.

\noindent \textbf{Baseline approach} \indent We use a sentence embedding suite to convert tweets to high-dimensional vectors that preserve semantic meaning in vector space. We used the Google Universal Sentence Encoder \cite{2018arXiv180311175C} as it offers good semantic representation regardless of the target task. We trained a deep neural network with two hidden layers on these sentence embeddings. While we achieved 83\% test accuracy on the Twitter test set, we noticed inaccurate and inconsistent predictions on long-form articles.

\begin{figure}[t]
    \centering
    \includegraphics[width=0.45\textwidth]{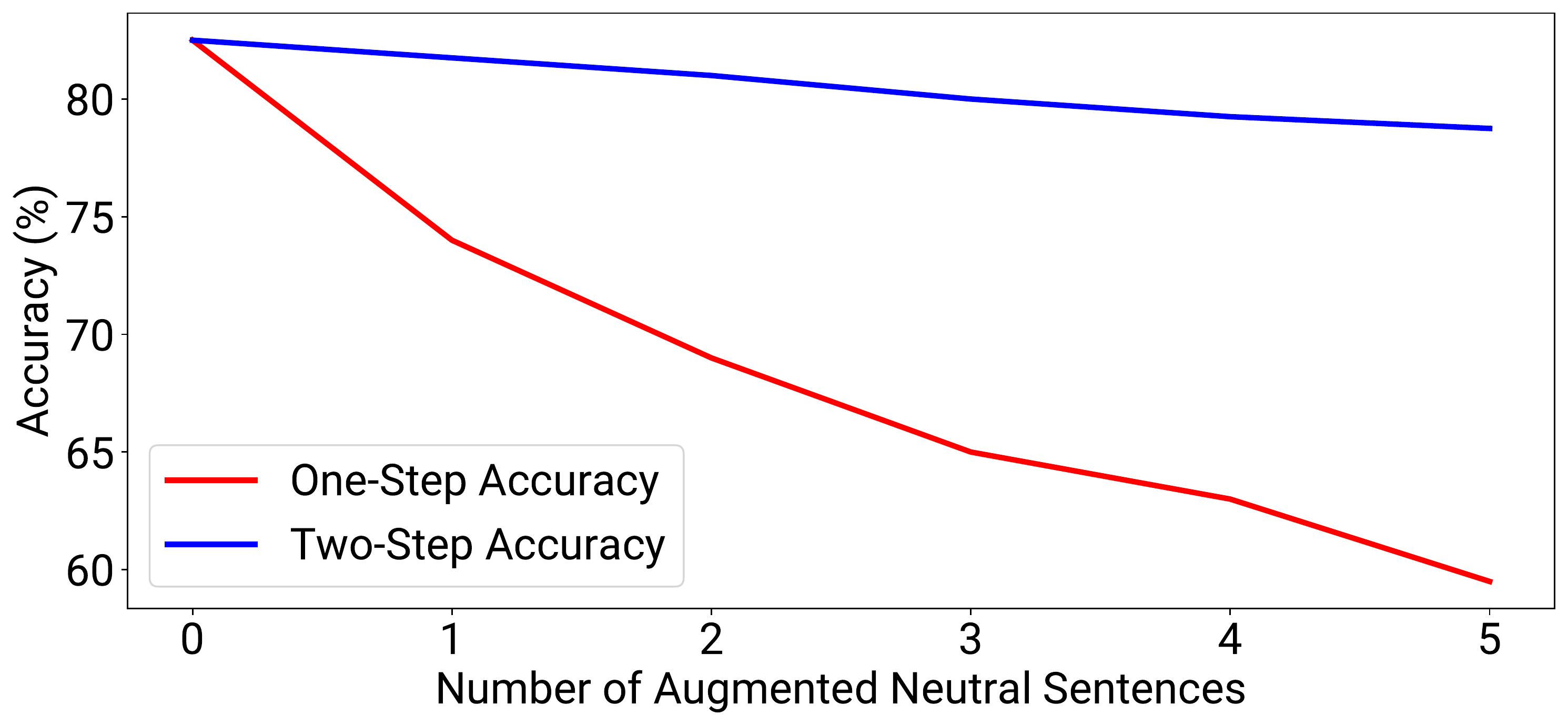}
    \caption{\small Degradation of accuracy after neutral sentence augmentation with One-Step vs. Two-Step classification approaches. The Two-Step method degrades gracefully relative to One-Step method as a result of removal of augmented sentences by neutral detector.}
    \label{degradation}
\end{figure}

\noindent \textbf{Opinion concentration} \indent We note that upon reading a number of long-form articles, a primary stylistic difference between these and tweets is the existence of neutral and apolitical sentences in the former medium. These sentences help article flow and cohesion, but also dilute the concentration of opinion compared to tweets. We hypothesize that this difference in opinion concentration is responsible for poor performance on long-form articles. We test this hypothesis by obtaining a set of neutral, apolitical sentences from the Twitter general stream and then augmenting them into the political test data. As demonstrated in Figure \ref{degradation}, accuracy decreases noticeably with the addition of augmented neutral sentences.

\begin{table}[t!]\label{tepc_results}
\fontsize{9.0pt}{9.0pt} \selectfont
\centering
\begin{tabular}{|l|l|l|l|l|l|}
\hline
Task                     & One-Step  & Two-Step \\ \hline
Twitter Political - Acc.          & 82.27\% & 82.42\%    \\ \hline
Twitter Crowdsourced - Acc.       & 86.00\%  & 86.00\%   \\ \hline
Twitter Crowdsourced - {$\rho$}   & 0.65  & 0.65    \\ \hline
Articles Crowdsourced - Acc.           & \textbf{66.67\%}  & \textbf{75.00\%}   \\ \hline
Articles Crowdsourced - {$\rho$}       & \textbf{0.52}  & \textbf{0.69}    \\ \hline
\end{tabular}
\caption{\small In bold are the experiments on long articles. Knowledge is transferred from learning on tweets at test time. All classifiers were DNN models with two hidden layers.}
\end{table}
\noindent \textbf{Neutral detector} \indent After identifying the dilution of opinion concentration as responsible for accuracy degradation on long-form articles, we propose the addition of a classifier to detect and remove neutral sentences. We train a second deep neural network on the sentence embeddings of 80,000 tweets sampled from the general Twitter stream as well as the political samples, obtaining a high 95.63\% accuracy. 

\noindent \noindent \textbf{Two-step classification scheme} \indent We propose a two-step classification scheme in order to improve prediction quality on long-form articles as demonstrated in Figure \ref{pipeline}. On any data passed to the system for inference, we first tokenize it into individual sentences. On each of these sentences, we use the neutral detector to mark and remove all neutral sentences. We then fuse the remaining sentences back together, aligning opinion concentration to that of tweets, and then use the main baseline classifier to predict polarity.

\section{Experiments}

\noindent \textbf{Datasets} \indent We tested our approach on a number of datasets. The first, Twitter Political, is a simple 20\% split of the obtained political tweet data consisting of 20,000 samples labeled based on authorship. We also manually selected a separate set of 50 tweets, as well as 24 articles from mainstream news outlets across the political spectrum, for which we collected crowdsourced annotations from 79 respondents. 

\noindent \textbf{Accuracy} \indent On long-form articles, the two-step method increased accuracy to 75\% from 66.7\%. However, as expected, the two-step method did not substantially improve accuracy on the Twitter datasets with as the opinion concentration remains the same due to the lack of neutral sentences.

\noindent \textbf{Spearman-Rho} \indent To verify prediction consistency, we computed the Spearman-rho rank correlation \cite{mcdonald_2015} against crowd opinions. Table \ref{tepc_results} shows that the proposed system ($\rho = 0.69)$ is far more consistent in assigning predictions with respect to crowdsourced predictions on articles than the baseline one-step method ($\rho = 0.52$).

\section{Conclusions \& Future Work}
We introduced a two-step classification method to detect polarity in text content without using metadata. By aligning opinion concentration using a neutral detector to remove apolitical sentences, our method performs well on tweets and long-form articles. Future work may involve exploring the problem of time shift, where predictions based on stale training data do not accurately represent positions on new issues. This reinforces the need for continuous model updates. Additionally, while we used random Twitter data to train the neutral detector and run the degradation experiment, it is desirable to test dilution by drawing neutral sentences that are more cohesive relative to the presented context, but this is somewhat difficult. We leave this open for future work. 

\fontsize{9.0pt}{10.0pt} \selectfont
\bibliographystyle{aaai}
\bibliography{refs1}
\textbf{Acknowledgement} The author thanks Prof. Kai-Wei Chang (UCLA) for comments and suggestions in writing this paper. 
\end{document}